\def\Year{1999}

\documentstyle[coin,twoside]{article}
\include{mssymb} 
\input{psfig}

\newcommand{\ma}[1]{$\mathrm{#1}$}
\newcommand{\mb}[1]{\[\mathrm{#1}\]}
\newcommand{\dstar}{\ma{d^{*}}}

\begin{document}

\title{AUTOMATICALLY SELECTING USEFUL PHRASES FOR DIALOGUE ACT TAGGING}

\maketitle

\author{Ken Samuel, Sandra Carberry, and K. Vijay-Shanker}

\affil{Computer and Information Sciences, University of Delaware, Newark, DE 19711, USA}

\begin{abstract}
We present an empirical investigation of various ways to {\em
automatically} identify phrases in a tagged corpus that are useful for
dialogue act tagging. We found that a new method (which measures a
phrase's deviation from an optimally-predictive phrase), enhanced with
a lexical filtering mechanism, produces significantly better cues than
manually-selected cue phrases, the exhaustive set of phrases in a
training corpus, and phrases chosen by traditional metrics, like
mutual information and information gain.
\end{abstract}

\section{INTRODUCTION}

Although machine learning approaches have achieved success in many
areas of Natural Language Processing, researchers have only recently
begun to investigate applying machine learning methods to
discourse-level problems~(Litman 1994, Andernach 1996,
Reithinger \& Klesen 1997, Wiebe et al. 1997, DiEugenio, Moore,
\& Paolucci 1997). An important task in
discourse understanding is to interpret an utterance's
\textbf{dialogue act}, which is a concise abstraction of the speaker's
intention; Figure~\ref{ex-das} presents a hypothetical dialogue that
has been labeled with dialogue acts. Recognizing dialogue acts is
critical for discourse-level understanding and can also be useful for
other applications, such as resolving ambiguity in speech recognition.
However, computing dialogue acts is a challenging task, because often
a dialogue act cannot be directly inferred from a literal
interpretation of an utterance.

\begin{figure*}[ht]
\centering
\begin{tabular}{|cclc|}
\multicolumn{1}{c}{\#} & Speaker & \multicolumn{1}{c}{Utterance} &
\multicolumn{1}{c}{Dialogue Act} \\
\hline
1 & John & Hello.                                          & Greet \\
2 & John & I'd like to meet with you on Tuesday at 2:00.   & Suggest \\
3 & Mary & That's no good for me,                          & Reject \\
4 & Mary & but I'm free at 3:00.                           & Suggest \\
5 & John & That sounds fine to me.                         & Accept \\
6 & John & I'll see you then.                              & Bye \\
\hline
\end{tabular}
\caption{A sample dialogue labeled with dialogue acts}
\label{ex-das}
\end{figure*}

We have investigated applying Transformation-Based Learning~(Brill
1995) to the task of computing dialogue acts. Transformation-Based
learning is a symbolic supervised machine learning method that
generates a sequence of rules. This method, which has not been applied
previously to discourse-level problems, has a number of attractive
characteristics for our task, such as its intuitive learned model and
its resistance to overfitting.~(Brill 1995)

Our machine learning algorithm makes use of several abstract features
extracted from utterances~(Samuel et al. 1998a). In particular, one of
the most effective features is the phrases\footnote{In this paper, the
term \textbf{phrase} refers to any sequence of one of more words that
may be found in a dialogue, such as ``by the way'' or ``how about
the''.} in an utterance that provide useful information for dialogue
act tagging, which we will call \textbf{dialogue act cues}. This paper
presents our investigation of methods for identifying dialogue act
cues.

We experimentally compared the effectiveness of various {\em
automatic} methods for selecting phrases, by applying them to a
\scshape VerbMobil \normalfont tagged corpus~(Reithinger \& Klesen
1997). This corpus consists of appointment-scheduling dialogues in
which each utterance has been manually labeled with one of eighteen
dialogue acts, such as Greet, Suggest, and Accept. Although we
understand that there may be problems\footnote{It may be relatively
difficult for human coders to label utterances with dialogue acts in a
consistent manner. Traditionally, intercoder reliability and intracoder
reliability have been significant problems for dialogue act
tagging.}\ma{^{,}}\footnote{There is substantial disagreement about
how to select an effective set of dialogue acts. Although several
researchers are currently addressing this problem (DRI 1997, MATE
1998, JDTWG 1999), the research community still lacks a standardized
set of dialogue acts.} with the dialogue acts in this corpus, we will
assume that they are correct, because these issues are beyond the
scope of this project. In any event, if another tagged corpus were to
become available, the methods presented here should be directly
applicable.

Our results showed that a {\em new} metric (which measures how far a
phrase deviates from an optimally-predictive phrase) enhanced with a
simple lexical filtering mechanism can select phrases that are more
effective for dialogue act tagging than phrases chosen by human
intuitive approaches or traditional metrics (like mutual information
and information gain).

\section{RELATED WORK}

Several researchers~(Cohen 1987, Fraser 1990, Grosz \& Sidner 1986,
Halliday \& Hasan 1976, Heeman, Byron, \& Allen 1998, Hirschberg \&
Litman 1993, Knott 1996, Marcu 1997, Reichman 1985, Schiffrin 1987,
Warner 1985, Zukerman \& Pearl 1986) identified \textbf{cue
phrases} that are useful for discourse processing, such as ``but'',
``so'', and ``by the way''. In most cases, their research focused on
selecting phrases that might be {\em generally} useful; however, we
have found that many of the phrases that appear to be useful for
our purposes were {\em not} included in the
previous literature. By analyzing the phrases and tags in a corpus,
automatic methods directly address three important factors:

\begin{enumerate}

\item The domain of discourse affects which phrases are useful. In the
appointment-scheduling dialogues of the \scshape VerbMobil \normalfont
corpus, phrases such as ``what time'' and ``I'm busy'' could be
effective.

\item The desired task of the system (dialogue act tagging, utterance segmentation,
etc.) has a significant impact. For dialogue act tagging, ``how
about'' and ``sounds great'' might serve as dialogue act cues.

\item The specific dialogue acts that we want to identify can affect
the usefulness of phrases. For example, one of the \scshape VerbMobil
\normalfont dialogue acts is Thank, so this motivates a need for phrases like
``thank you'' and ``thanks''.

\end{enumerate}

Intuitively, all of the phrases in the above examples seem to be
perfectly reasonable indicators of dialogue acts. However, to our
knowledge, no one has previously identified these phrases as cue
phrases. This leads us to suspect that the domain, task, and tags need
to be considered when selecting phrases.

\section{PHRASE-SELECTION METHODS}

The goal of our research is to devise a method that automatically
identifies dialogue act cues. This section discusses two baseline
approaches and several automatic methods, listed in
Figure~\ref{methods}.

\begin{figure}[t]
\centering
\begin{tabular}{rc}
\textbf{Method} & \textbf{Abbreviation} \\
\hline
Previous Literature                          & LIT \\
All Phrases                                  & ALL \\
\hline
Cooccurrences                                & COOC \\
Conditional Probability                      & CP \\
Entropy                                      & ENT \\
T Test                                       & TTEST \\
Mutual Information                           & MI \\
Selectional Preference Strength              & S \\
Information Gain                             & IG \\
Deviation                                    & D \\
Deviation Conditional Probability            & DCP \\
\hline
\end{tabular}
\caption{The various phrase-selection methods}
\label{methods}
\end{figure}

\subsection{Baseline Approaches}

We used two sets of phrases as baselines for comparison. 1)~The LIT
set consists of the 687 different cue phrases proposed in twelve
papers, dissertations, and books~(Cohen 1987, Fraser 1990, Grosz \&
Sidner 1986, Halliday \& Hasan 1976, Heeman, Byron, \& Allen 1998,
Hirschberg \& Litman 1993, Knott 1996, Marcu 1997, Reichman 1985,
Schiffrin 1987, Warner 1985, Zukerman \& Pearl 1986). 2)~The ALL set
represents an extreme approach, selecting {\em all} sequences of up to
three words found in a training corpus. Although this set is likely to
include all of the useful phrases, it also includes many extraneous
phrases, and we hypothesize that these irrelevant phrases can
overwhelm a machine learning algorithm.

\subsection{Automatic Methods}

Our general approach is to use {\em some metric} that estimates how
useful a phrase is for dialogue act tagging by analyzing the dialogue
acts of the utterances containing that phrase in a training corpus. In
this section, we will discuss the motivations and limitations of
several different metrics that we considered.

\subsubsection{Counting cooccurrences}

It is reasonable to expect that a dialogue act cue would cooccur
frequently with a specific dialogue act. For example, in the \scshape
VerbMobil
\normalfont corpus, the phrase ``see you'' is found in 106 utterances
that are labeled Bye, suggesting that ``see you'' is a dialogue act
cue. A straightforward way to rank phrases is to count how often each
phrase occurs in utterances labeled with each dialogue act. The
\textbf{cooccurence} method sorts phrases in decreasing order by their COOC
scores:
\nopagebreak
\mb{COOC(p) = \max_{d} \#(p\&d)}
\noindent
where p is a phrase, d is a dialogue act, and \#(x) is the frequency
(in the training corpus) of an event x. This metric maximizes over
dialogue acts in order to base the score on the best dialogue act for
the phrase.

\subsubsection{Considering dialogue act distribution}

The simple COOC metric does not take into account the {\em a priori}
distribution of dialogue acts. Unless each dialogue act is equally
likely, the most frequently-occurring dialogue acts will generate many
high-scoring phrases, even though that may be inappropriate. It might
be better to replace the joint frequency in COOC with the
\textbf{conditional probability} of a phrase given a dialogue act. The
conditional probability method sorts phrases in decreasing order by
their CP scores:

\mb{CP(p) = \max_{d} P(p|d)}
\noindent
where \ma{P(x|y)} is the probability of x given y.

Since COOC and CP maximize over dialogue acts, these scores only
account for one dialogue act for each phrase. But we might expect that
a dialogue act cue should cooccur frequently with a few dialogue acts and
infrequently with the others; a theoretically optimal dialogue act cue
would correlate {\em perfectly} with a single dialogue act, as
represented by the dashed line in Figure~\ref{optimal}. So it might be
worthwhile to consider how skewed the distribution of dialogue acts
cooccurring with a phrase is. This criterion is captured by the
\textbf{entropy} of the dialogue acts given a phrase. The entropy method\footnote{Our previous work~(Samuel et al. 1998b)
did not consider any metrics except for entropy.} sorts phrases in
increasing order by their ENT scores:

\begin{figure}[t]
\centerline{\psfig{figure=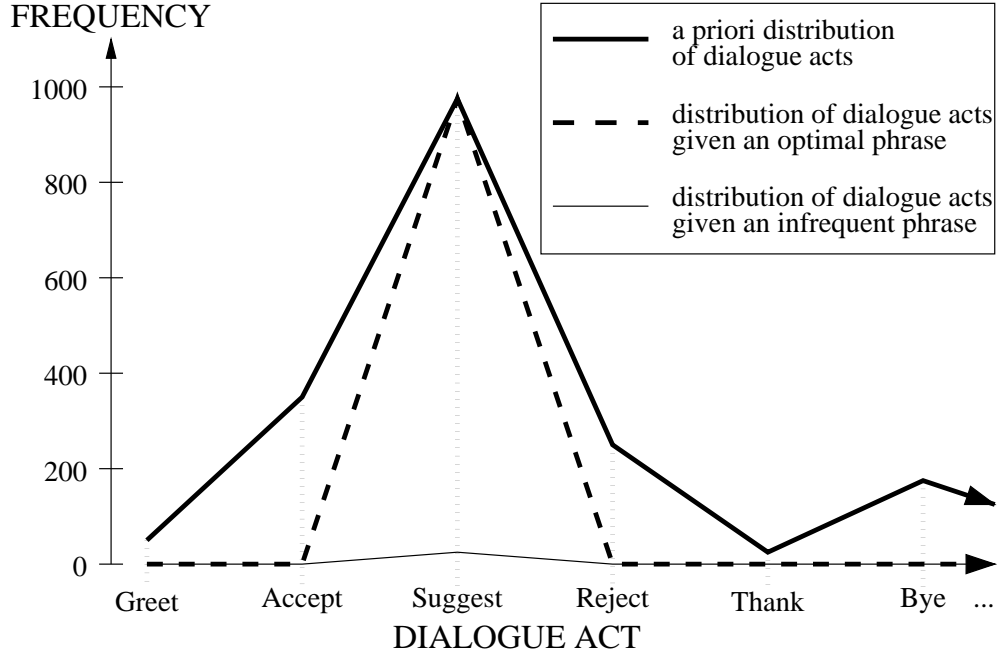}}
\caption{Dialogue act distributions}
\label{optimal}
\end{figure}

\mb{ENT(p) = -\sum_{d} P(d|p)\log_{2} P(d|p)}

However, like COOC, ENT does not account for the {\em a priori}
distribution of dialogue acts. Suppose the original dialogue act
distribution has relatively low entropy, and a phrase p is completely
independent\footnote{\ma{P(p|d)} is constant for all d.} of the
dialogue acts. Then ENT(p) is also relatively low (since the dialogue
act distribution is unaffected by the existence of p), incorrectly
signifying that p conveys useful information. To account for the a
priori dialogue act distribution, we examined four different metrics,
which are based on the Kullback-Liebler distance, mutual information,
the t test, and information gain.

The \textbf{selectional preference strength} method considers the
difference between the distribution of dialogue acts given a
particular phrase and the a priori distribution of dialogue acts to
estimate the amount of information a phrase carries about the dialogue
acts it cooccurs with.\footnote{Resnik~(1996) introduced the
selectional preference strength metric to measure how much information
about an argument of a verb would be lost by not taking into account
the verb itself.} This is a special case of the
\textbf{Kullback-Liebler distance} (also known as \textbf{relative entropy} or
\textbf{divergence}),\footnote{These
metrics are related to the \textbf{L1 norm} and
\textbf{information radius} (also known as \textbf{divergence
from the average}).} which measures how much information about a
dialogue act would be lost by failing to recognize a specific phrase.
The selectional preference strength method sorts phrases in decreasing
order by their S scores:

\mb{S(p) = \sum_{d} P(d|p)[\log_{2} P(d|p) - \log_{2} P(d)]}
\noindent
where \ma{P(x)} is the probability of x.

\textbf{Mutual information} has been used to measure the reduction of
uncertainty in one factor that results from the introduction of
another factor. We consider the mutual information between the
dialogue acts and a phrase, to compute the reduction of uncertainty in
an utterance's dialogue act when the utterance contains the phrase.
The mutual information method sorts phrases in decreasing order by
their MI scores:

\mb{MI(p) = P(p) \sum_{d} P(d|p)[\log_{2} P(d|p) - \log_{2} P(d)]}

\noindent
We note that, in this context, mutual information is closely related
to selectional preference strength.

The \textbf{t test} is used to measure the statistical difference between
two distributions. We ran a t test between the a priori distribution
of dialogue acts and the distribution of dialogue acts given a phrase.
The t test method sorts phrases in decreasing order by their TTEST
scores:

\mb{TTEST(p) = \#(\overline{p})
\sqrt{\sum_{d} \frac{D^{2}-D}{[D\#(p\&d)-\#(p)]^{2} + [D\#(d)-U]^{2}}}}
\noindent
where D is the number of different dialogue acts, U is the total
number of utterances, and \ma{\overline{p}} refers to the utterances
where p does not appear.

\textbf{Information gain} is typically utilized to estimate the usefulness of
a feature. For example, information gain has been used to determine
how to split a node in a decision tree, by considering the
distributions of data that fall along each branch. For our task, we
are testing for the existence of a phrase, so we use information gain
to measure the reduction in entropy of the dialogue acts resulting
from partitioning utterances based on whether or not they contain the
phrase. The information gain method sorts phrases in decreasing order
by their IG scores:

\mb{IG(p) = \sum_{d}[P(p)P(d\&p)log_{2}P(d\&p) +
P(\overline{p})P(d\&\overline{p})log_{2}P(d\&\overline{p}) - P(d)\log_{2}P(d)]}

\subsubsection{Measuring deviation from optimal}

In addition to adapting existing metrics, we also designed some new
metrics, which evaluate phrases based on their estimated effectiveness
in the hypothetical rule\footnote{In other words, the rule states that
phrase p appears in an utterance if and only if that utterance is
assigned the dialogue act d.} \texttt{p IFF d}. Recall that, if p is
an optimal dialogue act cue, it correlates perfectly with a single
dialogue act (like the dashed line in Figure~\ref{optimal}). And, if
\dstar~is that dialogue act,\footnote{In Figure~\ref{optimal},
\dstar~= Suggest.} then the rule \texttt{p IFF $\mathtt{d^{*}}$} is
{\em valid}. Therefore, this hypothetical dialogue act cue would be a
{\em perfect} indicator for the dialogue act $\mathtt{d^{*}}$. Our new
metrics measure how much each phrase deviates from this optimal design
by assigning a penalty point for each utterance where the rule fails.

There are two ways that the rule may fail. First, the rule may be
\textbf{unsound}, meaning that the left-to-right rule, \texttt{IF p THEN
$\mathtt{d^{*}}$}, applies incorrectly. For each utterance that
contains p but is not labeled \dstar, we assign a penalty point to the
phrase, for a total of \ma{\sum_{d \neq d^{*}} \#(p\&d)} points.
However, unsoundness alone is not sufficient. A phrase may produce a
perfect unsoundness score of 0, and yet still not be optimal. In the
extreme case, any phrase that appears only once in the training corpus
has an unsoundness score of 0.

One possible way to address this problem is to consider the thin line
in Figure~\ref{optimal}, which represents a phrase that only occurs in
25 utterances, where all 25 of those utterances are labeled Suggest.
Although this is certainly a useful phrase, since it is as sound as
the optimal phrase, we notice that there are 942 other Suggest
utterances, which do not include the phrase. We expect that another
equally-sound phrase that occurs more frequently should be ranked
higher, and so we are considering the case where the rule \texttt{p
IFF $\mathtt{d^{*}}$} is
\textbf{incomplete}, meaning that \texttt{IF $\mathtt{d^{*}}$ THEN p}
applies incorrectly. For each utterance that is labeled \dstar~but
does not contain p, we assign one penalty point to the phrase, for a
total of
\ma{\#(\overline{p}\&d^{*})} points.

It is unclear how to combine unsoundness and incompleteness in a
general metric. Certainly, when choosing between two equally sound
phrases, one would prefer the phrase that is more complete, and vice
versa. (See the next section for a qualitative analysis of empirical
results.) So, as an initial approach, we considered adding
incompleteness and unsoundness together. The
\textbf{deviation} method sorts phrases in decreasing order by their D
scores:

\mb{D(p) = \min_{d^{*}} [\#(\overline{p}\&d^{*})+\sum_{d \neq d^{*}} \#(p\&d)]}

\noindent
We minimize over dialogue acts, in order to base the score on the best
\dstar~for p.

Like COOC and ENT, the D metric does not account for the a priori
distribution of dialogue acts. So, we considered replacing the joint
frequencies in D with conditional probabilities. The \textbf{deviation
conditional probability} method sorts phrases in increasing order by
their DCP scores:

\mb{DCP(p) = \min_{d^{*}} [P(\overline{p}|d^{*})+\sum_{d \neq d^{*}} P(p|d)]}

\section{QUALITATIVE ANALYSIS}

We conducted some experiments to evaluate the merits of the various
metrics discussed in the last section. First, we used each metric to
order the phrases in the ALL set (all of the phrases in the \scshape
VerbMobil
\normalfont corpus). Then, we manually examined the highest-ranking
phrases to intuitively compare the methods. This qualitative analysis
immediately revealed some problems.

Several methods suffer from an undesirable bias based on frequency.
Many methods are susceptible to {\em infrequent} phrases; if a phrase
appears only once or twice in the corpus, we cannot really draw any
reliable conclusions about its usefulness. On the other hand, a number
of methods are biased toward phrases that appear {\em very frequently}
(such as ``the''). These phrases may be cooccurring frequently with
several (or all) of the dialogue acts, making them poor discriminators
of dialogue acts. To address a frequency bias, we might want to remove
any phrase with a frequency outside of some arbitrary range. However,
we believe it may be difficult (or even impossible) to find an
appropriate range, and so we would prefer to address this problem by
developing some {\em automatic} mechanism.

\begin{figure}[t]
\centering
\begin{tabular}{|r|cc|}
\hline
\textbf{Phrase} & \textbf{Unsoundness rank} & \textbf{Incompleteness rank} \\
\hline
we could meet &     3 &   1883 \\
how does the  &    14 &   2825 \\
yeah that     &    20 &   1174 \\
\hline	      	   
see you       &  8421 &     13 \\
hi            &  8285 &     20 \\
thanks        &  6813 &     26 \\
\hline
\end{tabular}
\caption{The tradeoff between unsoundness and incompleteness}
\label{ucp-vs-icp}
\end{figure}

In addition, we analyzed the tradeoffs between unsoundness and
incompleteness. Figure~\ref{ucp-vs-icp} lists some phrases that we
believe to be dialogue act cues, specifying how they would be ranked
based on unsoundness or incompleteness alone.\footnote{For this
figure, we used the conditional probability scoring method discussed
above.} For example, the phrase ``thanks'' occurs in eleven utterances
in the training corpus, and ten of these utterances are labeled with
the Thank dialogue act. As a result, it is assigned a very good
(though not perfect) unsoundness score. However, since every phrase
that occurs only once in the training corpus gets a perfect
unsoundness score, they all outrank ``thanks'' if only unsoundness is
considered. Alternatively, using incompleteness, ``thanks'' is ranked
\ma{26^{th}}, and the DCP method ranks it fifth.

The problem is that unsoundness is biased toward low-frequency phrases,
while incompleteness is biased toward high-frequency phrases. It is
not clear how to combine these two factors in order to balance their
biases. The D and DCP methods simply sum them, although we have also
considered weighting incompleteness and unsoundness in different ways.

Another potential problem is that, for several methods, many of the
highest-ranking phrases appear to address the same goal. For example,
all of the top eight ENT phrases (``how 'bout the'', ``'bout the'',
``okay how'', etc.) signal the Suggest dialogue act in basically the
same way. This is not surprising since, if one of these phrases
receives a good score, then they all should. However, we hypothesize
that the repetitions should be eliminated in order to produce
a more concise set of phrases, since this may increase the
effectiveness of the machine learning method in tagging dialogue acts.
Furthermore, if we want to select a predetermined number of phrases,
then a set with a wide variety of different phrases is probably more
useful than a set with many redundant phrases.

As a starting point, we can easily eliminate some of the redundant
phrases with a simple, lexical filtering mechanism, introduced in
Samuel et al.~(1998b). If one phrase contains another phrase as a
subsequence, and the second phrase is ranked higher, then the first
phrase is probably repetitious, and so it is unlikely to contribute
anything useful. For example, suppose the phrase ``see you'' is ranked
higher than ``will see you'', indicating that ``see you'' is more
informative. Since ``see you'' appears in every utterance where ``will
see you'' appears (and perhaps more), there is no good reason to keep
the phrase ``will see you''. The phrase ``see you'' has better
coverage and a better score, so it should always serve as a better
feature for dialogue act tagging. The \textbf{lexical filter} removes
a phrase if one of its subsequences is ranked higher.

\section{EXPERIMENTAL RESULTS}

We ran several experiments to compare the methods in
Figure~\ref{methods} on the task of labeling utterances with dialogue
acts. For all of these experiments, we applied Transformation-Based
Learning~(Brill 1995) using three classes of features that we
have experimentally found to be particularly effective:

\begin{itemize}

\item Applying one of the methods described above to rank the phrases,
the system used the best-rated phrases as features of utterances.

\item An attractive characteristic of Transformation-Based Learning is that it generates
preliminary tags during training. These tags can then be used as
features to further refine the learned model. Ramshaw and
Marcus~(1994) referred to this as ``leveraged learning''. So, to help
determine the dialogue act of a given utterance, our system used the
preliminary dialogue act assigned to the {\em preceding} utterance as
a feature.

\item Our system utilized a ``change-of-speaker'' feature that represented information about
the speaker of a given utterance. This boolean feature is True for an
utterance if the speaker of that utterance differs from the speaker of
the preceding utterance, and False otherwise.

\end{itemize}

An effective heuristic is to cluster certain words into semantic
classes, which can collapse several dialogue act cues into a single
dialogue act cue. For example, in the appointment-scheduling corpora,
there is a strong correlation between utterances that mention weekdays
and the Suggest dialogue act, but to express this fact, it is
necessary to consider five separate dialogue act cues, such as: ``on
Monday the'', ``on Tuesday the'', ``on Wednesday the'', ``on Thursday
the'', and ``on Friday the''. However, if the five weekdays are
combined under one label, ``\$weekday\$'', then the same information
can be captured by a single dialogue act cue that has five times as
much data supporting it. The experiments presented in this paper use
the following semantic clusters: ``\$weekday\$'', ``\$month\$'',
``\$number\$'', ``\$ordinal-number\$'', and ``\$proper-name\$''.

All of our experimental results were derived from a set of held-out
data (328 utterances), which was completely disjoint from the training
data (2701 utterances) that we used to select phrases.

\subsection{The Cutoff Points}

\begin{figure}[t]
\centerline{\psfig{figure=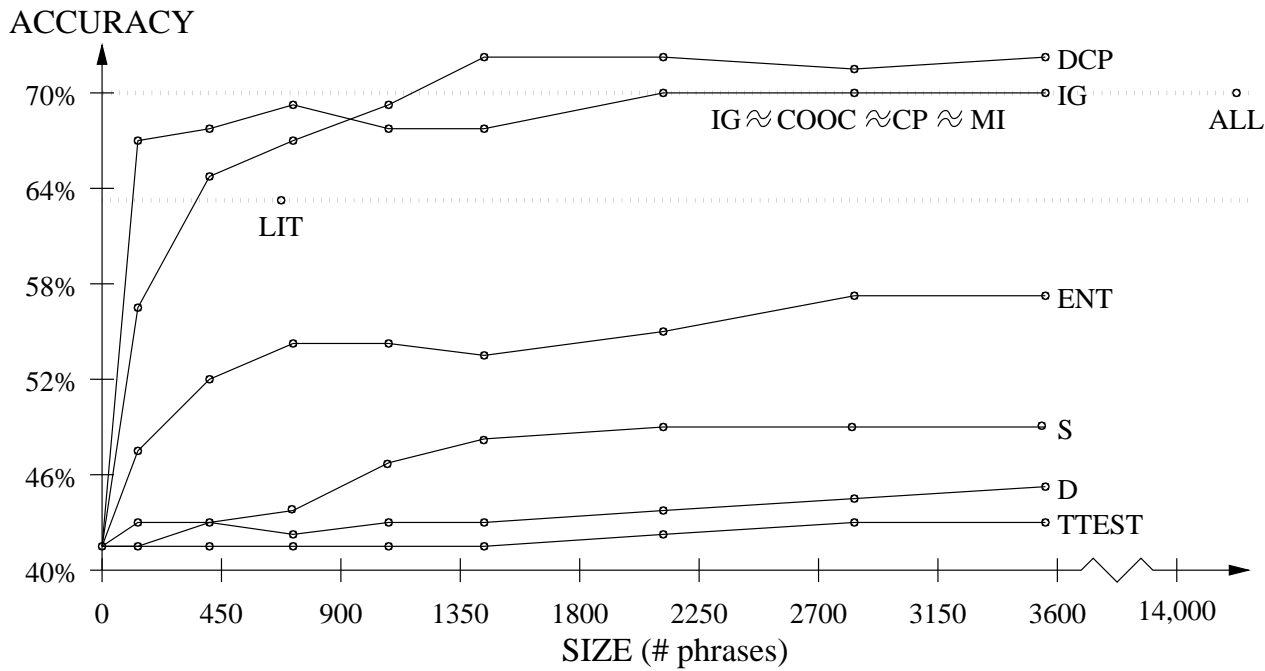}}
\caption{Size versus accuracy without the filter}
\label{cutoff-lexical-results}
\end{figure}

Since the methods are supposed to rank dialogue act cues higher than
other phrases, we should be able to separate the dialogue act cues
from the other phrases. To test this, we applied various cutoff points
to each method to determine how many lower-ranking phrases may be
removed before accuracy begins to decrease. We wanted to investigate
the cutoff points in isolation, so the lexical filter was not used in
this set of experiments. Figure~\ref{cutoff-lexical-results} presents
the accuracy of each method as a function of the number of phrases
used. The ALL and LIT sets are also included in the figure, for
comparison. (For clarity, COOC, CP, and MI are not shown in the
figure, because their curves are similar to IG's curve.)

Four methods, TTEST, D, S, and ENT, produced accuracies
significantly\footnote{In all of the experiments in this paper, the
differences were analyzed for statistical significance with the t
test~(Levine 1981) or the Tukey ``honest significant differences''
test, which is an extension of the t test that is appropriate for
comparing more than two distributions.~(Masterson 1997)} below LIT
when 25\% (3558) of the 14,231 phrases were selected. This implies
that many dialogue act cues were ranked in the bottom 75\% by these
methods, suggesting that there may be a problem with these phrase
orderings. On the other hand, for four methods, IG, COOC, CP, and MI,
we could remove more than 13,000 phrases without significantly
affecting the accuracy. These methods also produced significantly
higher accuracy scores than the LIT set. Therefore, automatic methods
can select phrases that are better for dialogue act tagging than the
cue phrases found in the literature.

However, DCP was the only method that produced a significant {\em
rise} in accuracy over ALL. With cutoff points from 10\% (1423) to
25\% (3558), DCP's accuracy was significantly higher than ALL's
accuracy. As we hypothesized above, it appears that the irrelevant
phrases in ALL limit the accuracy of the machine learning method. And
we expect that this effect would be more pronounced for a larger
training corpus (with more phrases) or another machine learning method
(that is more susceptible to irrelevant features).

However, for cutoff points of 5\% (712) and lower, DCP is
significantly worse than ALL. We believe that this is because DCP is
susceptible to repetitive phrases. Since DCP assigns high scores to
many redundant phrases, we require a relatively large set of phrases
in order to capture the full variety of dialogue act cues. This is
precisely the problem that the lexical filter was designed to address.

\subsection{The Lexical Filter}

Our next set of experiments tested the lexical filter. If the phrases
are ordered properly, then the filter should eliminate some redundant
phrases without compromising accuracy in labeling dialogue acts.
First, we ordered the phrases with each method and applied the filter.
Then, we used various cutoff points to select the top-ranked phrases
for training and testing our dialogue act tagger.

\begin{figure}[t]
\centerline{\psfig{figure=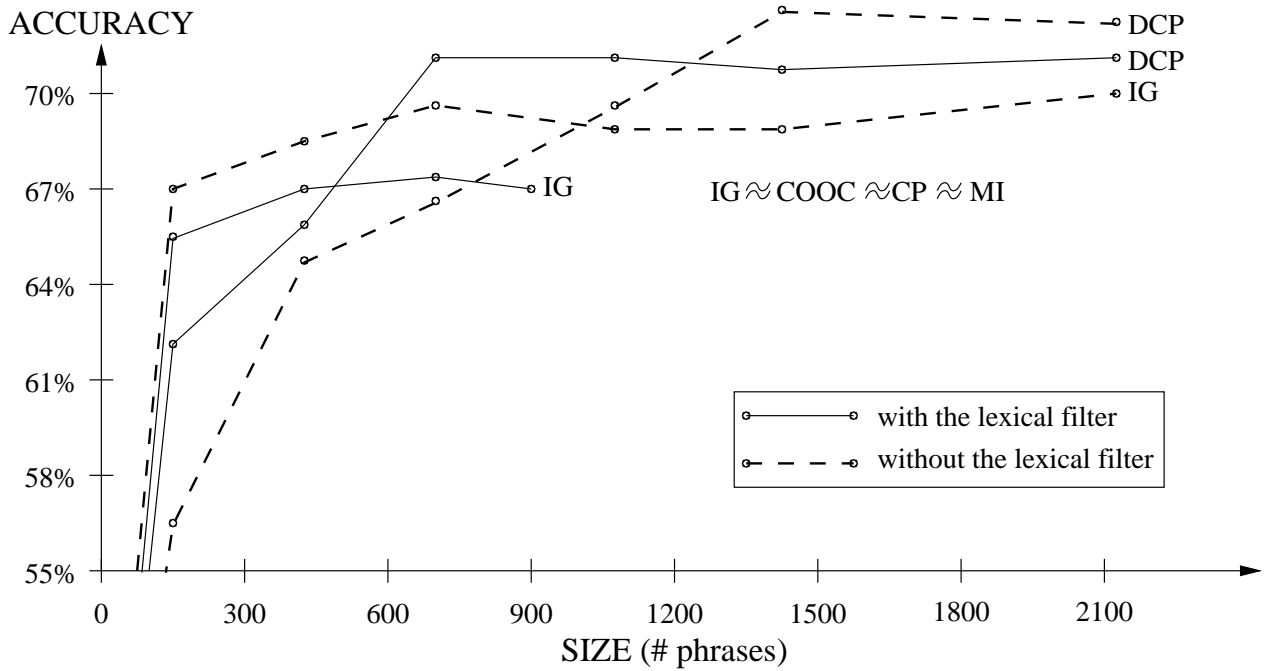}}
\caption{Size and filter versus accuracy}
\label{old-filter}
\end{figure}

This produced some unexpected results, as shown in
Figure~\ref{old-filter}.\footnote{In this figure, three methods, COOC,
CP, and MI, are again omitted for clarity, because their results were
similar to IG's results. Also, with the lexical filter, these four
curves don't extend beyond 900 phrases, because the lexical filter
removes 94\% (13,342-13,347) of the phrases in each case.} We found
that, in some cases, the filter significantly {\em decreased} the
accuracy on the dialogue act tagging task. Since we expected the
filter to remove phrases {\em without} compromising accuracy, this
result prompted us to analyze the filter's design more carefully.

We now believe that it is important for the filter to consider {\em
why} a phrase is being selected. For example, while the phrase ``hi''
tends to signal the Greet dialogue act, an utterance with ``hi I'' is
more likely to be an Init. Our filter would erroneously remove the
phrase ``hi I'', losing some relevant information. So, we modified our
filter to follow this new rule:

\begin{center}
\begin{tabular}{l}
IF a phrase p has a subsequence p' that is ranked higher \\
{\em AND both p and p' were selected for the same dialogue act} \\
THEN remove p \\
\end{tabular}
\end{center}

The second condition requires further explanation. For the COOC, CP,
D, and DCP methods, the metrics
maximize (or minimize) over dialogue acts. So, for a given phrase, we
determine which dialogue act is producing the maximum (or minimum)
value, and define that to be the dialogue act referred to in the
second condition. For the other methods, the metrics sum over dialogue
acts. In these
cases, we follow Resnik~(1996) by selecting the dialogue act that
produces the greatest contribution to the sum.\footnote{For TTEST, the sum is not located on the outside of the
formula. However, since the square root function is monotonic and
\ma{\#(\overline{p})} is constant for a given phrase, we can use the same
approach for selecting a dialogue act with the TTEST method.}

\begin{figure}[t]
\centering
\begin{tabular}{|cr|}
\hline
\textbf{Method} & \multicolumn{1}{c|}{\textbf{Size}} \\
\hline
LIT   &    687 \\
\hline
COOC  &  3994 \\
MI    &  4291 \\
IG    &  5202 \\
CP    &  5515 \\
DCP   &  8509 \\
ENT   &  9610 \\
S     &  9635 \\
TTEST & 10,189 \\
D     & 11,007 \\
\hline
ALL   & 14,231 \\
\hline
\end{tabular}
\caption{Experimental results with the modified filter}
\label{filter-results}
\end{figure}

\begin{figure}[t]
\centerline{\psfig{figure=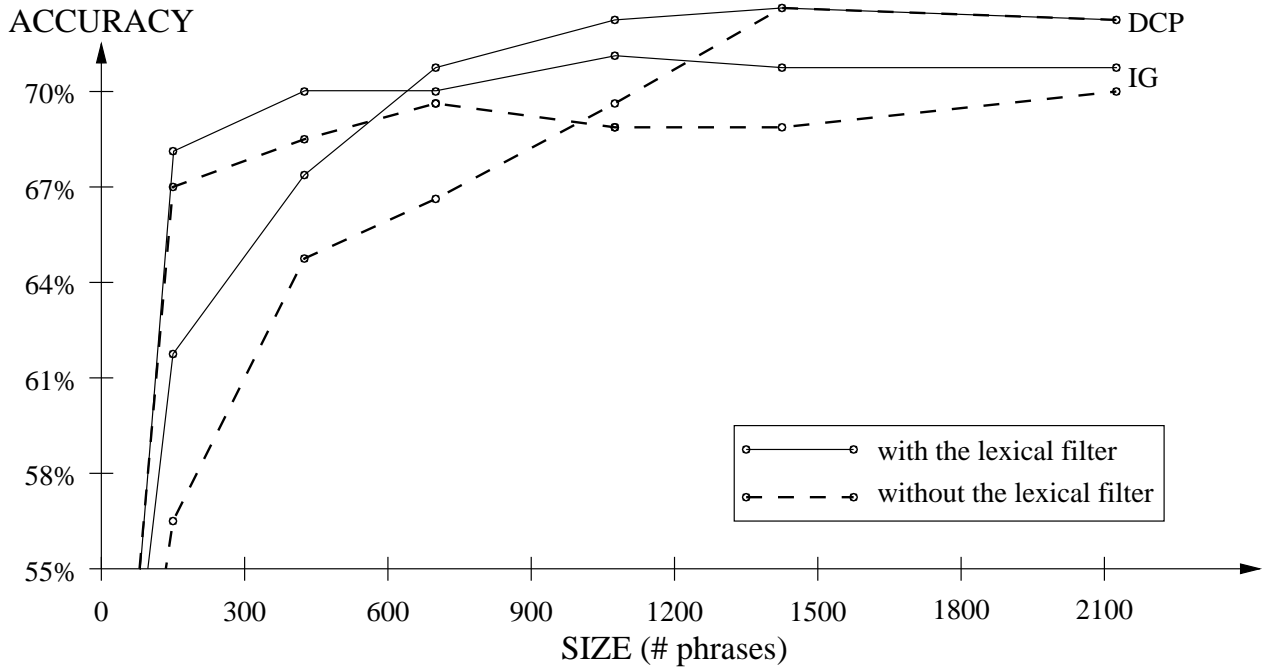}}
\caption{Size and modified filter versus accuracy}
\label{modified-filter}
\end{figure}

The effect of this modified filter varies dramatically, removing 23\%
(3224) to 72\% (10,237) of the 14,231 phrases, as shown in
Figure~\ref{filter-results}. However, Figure~\ref{modified-filter}
shows that, as expected, using the filter does not cause the accuracy
to decrease. In addition, it allows the system to maintain a high
accuracy with fewer phrases. In particular, DCP's accuracy is
significantly higher than ALL's accuracy when using only 5\% (712) of
the phrases in ALL. This suggests that the filter is effectively
removing redundant phrases, to produce a more parsimonious set of
phrases.

\section{DISCUSSION}

This paper presented an investigation of various methods for selecting
useful phrases. We argued that the traditional method of selecting
phrases, in which a human researcher analyzes discourse and chooses
general cue phrases by intuition, could miss useful phrases. To
address this problem, we introduced {\em automatic} methods that use a
tagged training corpus to select phrases, and our experimental results
demonstrated that these methods can outperform the manual approach.
Another advantage of automatic methods is that they can be easily
transferred to another tagged corpus.

Our experiments also showed that the effectiveness of different
methods on the dialogue act tagging task varied significantly, when
using relatively small sets of phrases. The method that used our new
metric, DCP, produced significantly higher accuracy scores than any of
the baselines or traditional metrics that we analyzed. In addition, we
hypothesized that repetitive phrases should be eliminated in order to
produce a more concise set of phrases. Our experimental results showed
that our modified lexical filter can eliminate many redundant phrases
without compromising accuracy, enabling the system to label dialogue
acts effectively using only 5\% of the phrases.

There are a number of research areas that we would like to investigate
in the future, including the following: We intend to experiment with
different weightings of unsoundness and incompleteness in the DCP
metric; we believe that the simple lexical filter presented in this
paper can be enhanced to improve it; we would like to study the merits
of enforcing frequency thresholds for methods that have a frequency
bias; for the semantic-clustering technique, we selected the clusters
of words by hand, but it would be interesting to see how a taxonomy,
such as WordNet, could be used to automate this process; since all of
the experiments in this paper were run on a single corpus, in order to
show that these results may generalize to other tasks and domains, it
would be necessary to run the experiments on different corpora.

\section{ACKNOWLEDGMENTS}

The members of the \textsc{VerbMobil} research group at DFKI in
Germany, including Norbert Reithinger, Jan Alexandersson, and
Elisabeth Maier, generously granted us access to the
\textsc{VerbMobil} corpora. This work was partially supported by the
NSF Grant \#GER-9354869.

\end{document}